\documentclass{article}

\usepackage{PRIMEarxiv}

\usepackage[utf8]{inputenc} 
\usepackage[T1]{fontenc}    
\usepackage{hyperref}       
\usepackage{url}            
\usepackage{booktabs}       
\usepackage{amsfonts}       
\usepackage{nicefrac}       
\usepackage{microtype}      
\usepackage{lipsum}
\usepackage{fancyhdr}       
\usepackage{graphicx}       
\graphicspath{{media/}}     
\usepackage{times}
\usepackage{soul}
\usepackage{amsmath}
\usepackage{amsthm}
\usepackage{booktabs}
\usepackage{algorithm}
\usepackage{algorithmic}
\usepackage{longtable}
\usepackage{multirow}
\usepackage{xcolor}
\usepackage{dblfloatfix}
\usepackage{xspace}
\title{A Comprehensive Review on Deep Supervision: Theories and Applications}

\author{
  Renjie Li\\
  School of Information and Communication Technology\\
  University of Tasmania, Hobart \\
   \And
  Xinyi Wang\\
  School of Information and Communication Technology\\
  University of Tasmania, Hobart \\
     \And
  Guan Huang\\
  School of Information and Communication Technology\\
  University of Tasmania, Hobart \\
     \And
  Wenli Yang\\
  School of Information and Communication Technology\\
  University of Tasmania, Hobart \\
     \And
  Kaining Zhang\\
  School of Information and Communication Technology\\
  University of Tasmania, Hobart \\
     \And
  Xiaotong Gu\\
  School of Information and Communication Technology\\
  University of Tasmania, Hobart \\
     \And
  Son N. Tran\\
  School of Information and Communication Technology\\
  University of Tasmania, Hobart \\
     \And
  Saurabh Garg\\
  School of Information and Communication Technology\\
  University of Tasmania, Hobart \\
     \And
  Jane Alty\\
  Wicking Dementia Research and Education Centre\\
  University of Tasmania, Hobart \\
     \And
  Quan Bai\\
  School of Information and Communication Technology\\
  University of Tasmania, Hobart \\
}

\begin{document}
\maketitle

\begin{abstract}
Deep supervision, or known as `intermediate supervision' or `auxiliary supervision', is to add supervision at hidden layers of a neural network. This technique has been increasingly applied in deep neural network learning systems for various computer vision applications recently. There is a consensus that deep supervision helps improve neural network performance by alleviating the gradient vanishing problem, as one of the many strengths of deep supervision. Besides, in different computer vision applications, deep supervision can be applied in different ways. How to make the most use of deep supervision to improve network performance in different applications has not been thoroughly investigated. In this paper, we provide a comprehensive in-depth review of deep supervision in both theories and applications. We propose a new classification of different deep supervision networks, and discuss advantages and limitations of current deep supervision networks in computer vision applications.
\end{abstract}


\section{Introduction}


In recent years, deep neural networks (DNNs) have gained considerable attention in the computer vision field. Thanks to the capability of DNNs (particularly the Convolutional Neural Networks (CNNs)) to learn deep and meaningful features from large amounts of image and video datasets, it brings success to various computer vision applications including image classification, object detection, image segmentation, keypoint detection, and super-resolution. Researchers have been focused on designing different CNN structures to fit various computer vision applications. VGG \cite{simonyan2014very} and ResNet \cite{he2016deep} achieve success in image classification, and also can be used as backbone networks in object detection frameworks such as YOLO \cite{redmon2016you} and Faster R-CNN \cite{ren2015faster}. Fully Convolutional Network (FCN) \cite{long2015fully} achieves success in semantic segmentation, while U-Net \cite{ronneberger2015u} lays the foundation for medical image segmentation. Stacked hourglass network \cite{newell2016stacked} achieves success on human pose estimation. Different CNN structures are being developed to improve performance in various computer vision applications. However, designing different CNN structures is not the only way to improve performance. How to improve the network learning efficiency is also a critical point to achieve a good performance. 

Deep supervision is an idea to add supervision at hidden layers of the network, and the loss from such supervision is combined together in the objective function for learning the weights~\cite{lee2015deeply}. Unlike designing different CNN structures, deep supervision is a technique that can be applied to current CNN structures to improve the performance. There is a consensus that deep supervision can improve the neural network performance because it can alleviate the gradient vanishing problem in order to make the network very deep, and help networks learn more discriminative features used in different ways. Wang et al. \cite{wang2015training} proposed a network structure with deep supervision being applied at different depth levels of the network for image classification. In the face keypoint localisation domain, Bulat and Tzimiropoulos \cite{bulat2016convolutional} proposed a Convolutional Aggregation of Local Evidence (CALE) network, with deep supervision being applied to guide the network to learn face keypoint positions from the facial part detected, with more than 50\% gain in localisation accuracy. In medical image segmentation, Dou et al. \cite{dou20163d} proposed a 3D deeply supervised network for liver segmentation based on CT volumes where deep supervision is employed at different levels of the network. The results also showed that deep supervision helps improve performance. In addition to the performance improvement, deep supervision also brings other benefits to neural network learning systems, such as providing more transparent insight into the training process and increasing the convergence speed in the training process \cite{lee2015deeply}.    


Given the advancement in deep supervision networks and their promising performance in computer vision tasks, there is a need for a comprehensive review synthesising the current theories and applications to direct future research. The focus of most research studies has been on either just applying the deep supervision in existing CNNs or discussing the deep supervision in a specific computer vision task. Most researchers have not given the overall dimensions involved in deep supervision such as the in-depth reason why deep supervision takes effect and how to use deep supervision in the best way in different computer vision applications. Therefore, it would be useful for researchers to have an in-depth summary of the research into deep supervision networks to better understand deep supervision, including associated risks, and to support the use of deep supervision across a range of applications. To the best of our knowledge, no related deep supervision review paper exists. We have undertaken a comprehensive review of the literature on deep supervision networks, including both theories and computer vision applications. This paper makes the following concrete research contributions: 
\begin{itemize}
\item Create a novel way to classify deep supervision networks;
\item Make a comprehensive review of how different computer vision applications can be benefited from deep supervision;
\item Clearly identify the advantages and risks of deep supervision when being applied.
\end{itemize}




This paper is structured as follows: Section II introduces the background of deep supervision. Section III presents a new classification of deep supervision networks. Section IV introduces learning strategies that are used in deep supervision networks. Section V presents how deep supervision has been used in the key computer vision tasks. Section VI presents the conclusions and future directions.

\section{Background}
In this section, we outline the background of deep supervision, and clarify two cases which are not scoped within this review paper.

\subsection{Overview}
Deep supervision was first applied by Lee et al.~\cite{lee2015deeply} when they proposed Deeply Supervised Net (DSN) for an image classification task in 2015. In DSN, hidden layers are supervised by the main task label to accelerate the convergence speed in the training. This study demonstrated that deep supervision improves the performance in image classification tasks due to its ability to mediate the vanishing gradient issue and the slow training convergence issue.

\begin{figure}[ht]
\centering
\includegraphics[width=0.50\textwidth]{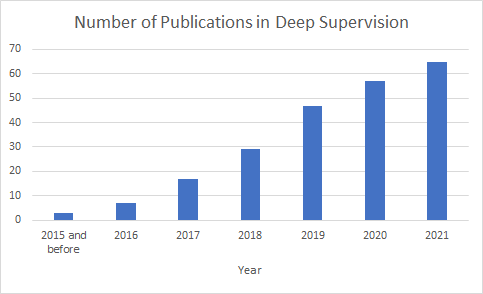}
\caption{The annual number of publications on deep supervision has increased between 2016 and 2021.  (This data is retrieved by using the following advanced search keywords in Scopus and IEEE explorer database: 'deep supervision' AND 'deeply supervised net' AND 'intermediate supervision'.)}
\label{fig:PublicationNumber}
\end{figure}

Encouraged by the success of deep supervision in image classification task, there has been an increasing number of publications starting to apply deep supervision in CNNs to improve performance in various computer vision applications as shown in figure \ref{fig:PublicationNumber}, such as image segmentation \cite{dou20163d,asl2018alzheimer}, object detection \cite{li2019real,RN48}, keypoint detection \cite{wu2019lightweight} and image super-resolution \cite{kim2016deeply}.

\subsection{Different shades of deep supervision}
Deep supervision talks about adding extra supervision at the hidden layers of the neural network to improve the whole neural network performance \cite{lee2015deeply}. Since supervision is employed at the hidden layer of the neural network, and in most cases, such supervision serves as a supporting role, that's why in some works, deep supervision is also called `intermediate supervision' or `auxiliary supervision'. In fact, they all refer to the same thing - deep supervision. In the following paper, we refer to both `intermediate supervision' and `auxiliary supervision' as `deep supervision' for simplicity.

\subsection{Deep supervision vs multi-tasks training}
By introducing deep supervision to the hidden layers of the whole neural network, more loss is integrated together into the objective function of the learning process to update weights. Although extra loss is introduced to the main objective function, deep supervision is different from those which includes multiple loss from the final layer. For instance, in the work of YOLOv1 \cite{redmon2016you}, multiple loss was defined for different purposes (e.g. bounding box coordinates, class probabilities) and applied at the final layer of the network. This is not an application of deep supervision. Besides, multi-task problem supervised by multiple labels in a single neural network is also not considered as deep supervision applications. For example, Ege et al. \cite{ege2017simultaneous} proposed a simultaneous CNN to classify both food categories and calories. This is a dual-task classification problem solved by a single CNN supervised by multiple labels, not an application of deep supervision. Both types of works are not included in this review paper.

\subsection{Deep supervision vs multi-phase training}
Deep supervision is also different from multi-phase training. Hinton et al.~\cite{hinton2006fast} proposed deep belief network, a network consisting of different Restricted Botzmann Machine (RBM). The deep belief network updated the weights layer by layer through unsupervised learning. Specifically, input data and the first hidden layer were regarded as an RBM, and the weights in this RBM were being trained and fixed before training for the next RBM. There are also some multi-phase training works in computer vision field. Honda et al.~\cite{honda2018enhanced} proposed a two-stage multi-person pose estimation network. The network needs to be trained for two phases. In the first phase, the input was an image and the outputs were several region-of-interests (ROIs). In the second phase, ROIs were input and the outputs were heatmaps of each keypoint of human body. The two stages were trained separately and using different labels. The main difference between multi-phase training and deep supervision is that multi-phase training treats the network as different phases, while deep supervision occurs in one end-to-end network (only one phase). In multi-phase training, network is separated into different phases, and the subsequent phase will not start training until the previous phase completes training. In deep supervision, although the weights for a certain deep supervision can be fixed, still the training happens in one end-to-end network. Multi-phase training is not scoped within this review paper.


\section{Classification of Deep Supervision Networks}
In this section, we categorise deep supervision network into hidden layer deep supervision (HLDS), different branches deep supervision (DBDS) and deep supervision post encoding (DSPE). In the following, we explain the details of each category.

\subsection{Category 1 - Hidden Layer Deep Supervision (HLDS)}
\begin{figure}[ht]
\centering
\includegraphics[width=0.27\textwidth]{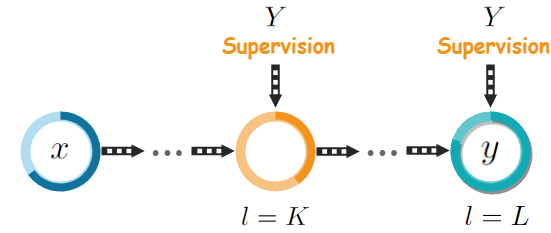}
\caption{Category 1 HLDS: deep supervision is simply employed in some hidden layers of the neural network.}
\label{fig:type1}
\end{figure}

In HLDS, deep supervision is simply employed at some hidden layers of DNNs. For example in Figure~\ref{fig:type1}, deep supervision is employed at the $K^{th}$ layer, so that the features learned at the $K^{th}$ layer are more discriminative than other layers.

In the forward propagation phase, the introduction of deep supervision does not change the way of forward information propagation (see Equation~\ref{eq:type1_forward}).

\begin{align}
& y = f^{(l=L)}(...f^{(l=K)}(...f^{(l=1)}(x; \theta_{1})...; \theta_{K})...;\theta_{L}) \label{eq:type1_forward}
\end{align}
where $l$ is a layer number and $L>K$, $x$ is the input, $y$ is the output and $f^{(l=l)}( ; \theta_{l})$ stands for a function with parameters of $\theta_{l}$ at the $l^{th}$ layer, deep supervision is employed at the $K^{th}$ layer and the final layer is the $L^{th}$ layer.



The loss function is defined in Equation~\ref{eq:type1_loss}. Total loss is a combination of the final output layer loss ($\mathcal{L}_L$) and intermediate supervision loss ($\mathcal{L}_K$ ). In most cases, there is a hyper-parameter ($\alpha$) to control the balance between the intermediate supervision loss and the final output layer loss.

\begin{align}
& \mathcal{L}=\mathcal{L}_L + \alpha \times \mathcal{L}_K \label{eq:type1_loss}
\end{align}

In the backward propagation phase, the parameters of layer $L$ to layer $K+1$ are learned and updated by minimising the final output layer loss, while the parameters of layer $K$ to layer $1$ are learned and updated by minimising the total loss (Equation~\ref{eq:type1_back}). Generally, if $L$ is very large (i.e. the neural network is very deep), the gradient vanishing problem is prone to occur. With the help of deep supervision at the $K^{th}$ layer, when weights are updated till layer $K$, a new loss-based gradient is introduced so that it smooths the gradient in the back propagation, and further, solves the gradient vanishing problem to a large extent.

\begin{align}
& \theta^{+}_{l}=\left\{
\begin{aligned}
\theta_{l}-\eta \times \frac{\partial \mathcal{L}_{L}}{\partial \theta_{l}} & , l=K+1,...,L.\\
\theta_{l}-\eta \times (\frac{\partial \mathcal{L}_{L}}{\partial \theta_{l}}+ \frac{\partial \mathcal{L}_{K}}{\partial \theta_{l}}) & , l=1,...,K.
\end{aligned}
\right. \label{eq:type1_back}
\end{align}
where $\theta_{l}$ is the parameter at the $l^{th}$ layer, $\eta$ is the learning rate, $\theta^{+}_{l}$ is the updated parameter at the $l^{th}$ layer, $\mathcal{L}_{L}$ is the loss at the $L^{th}$ layer (the final layer) and $\mathcal{L}_{K}$ is the loss at the $K^{th}$ layer (the layer where deep supervision is applied.

Deep supervision can be added to different hidden layers to make the gradient flow smoothly during backward propagation. The number of deep supervision being employed is a critical point. If deep supervision is included in too many hidden layers, the network will converge too quickly and lose the generalization capability. In contrast, too less deep supervision added may not take full advantage of deep supervision to solve the gradient vanishing problem. Where to add deep supervision along a whole neural network is also a critical point. In HLDS, it is optimal to add deep supervision to hidden layers where the gradient is close to 0 in the backward propagation. More details are discussed in the Learning section.

\subsection{Category 2 - Different Branches Deep Supervision (DBDS)}
\begin{figure}[ht]
\centering
\includegraphics[width=0.5\textwidth]{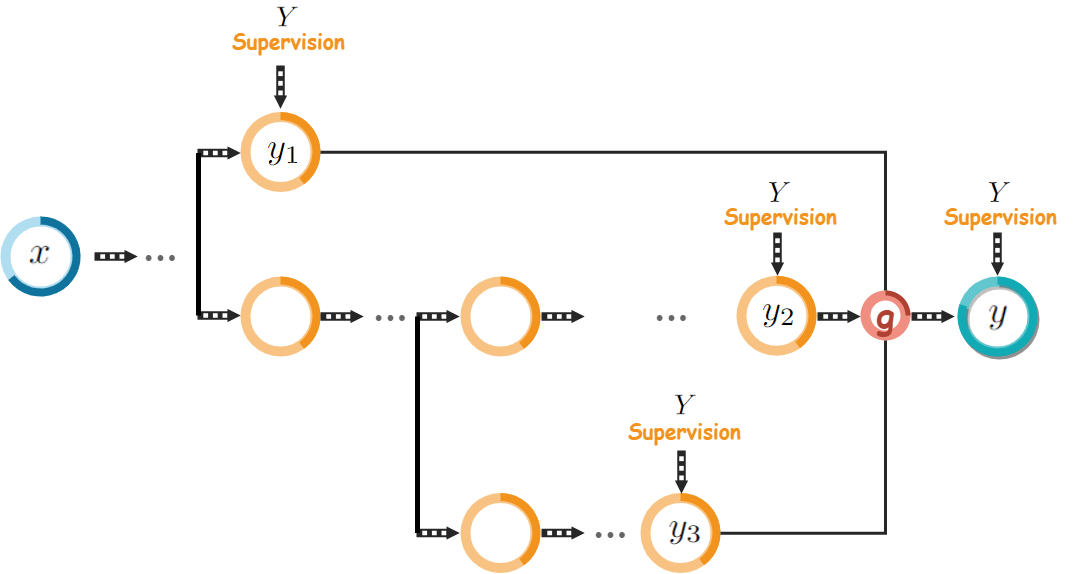}
\caption{Category 2 DBDS: deep supervision is employed at different branches and different depth layers of the neural network.}
\label{fig:type2}
\end{figure}

In DBDS, deep supervision is employed at different branches and different depth layers of the neural network. In Figure~\ref{fig:type2}, three deep supervision is employed separately at different branches and depth layers: a shallower layer at the $1^{st}$ branch, a deeper layer at the $2^{nd}$ branch and a middle layer at the $3^{rd}$ branch. Then, the three deeply supervised layers' features are integrated together by using operator $g()$ (operator $g()$ can be concatenation, sum-up, etc).

In the forward propagation phase, the information spread along different branches, and the deeply supervised layers are integrated together for the final output. Equation~\ref{eq:type2_forward1} and~\ref{eq:type2_forward2} describe the forward propagation phase of Figure~\ref{fig:type2}.

\begin{align}
& y_{i}=f^{y_{i}}(x_{i}; \theta_{i}) \label{eq:type2_forward1}, i=1, 2, 3 \\
& y=g(y_{1}, y_{2}, ... y{i}), i=1, 2, 3 \label{eq:type2_forward2}
\end{align}
where $i$ is a branch number, $x_{i}$ is the input to get $y_{i}$, $f^{yi}(; \theta_{i})$ is a function with parameters of $\theta_{i}$ that outputs $y_{i}$, $g()$ is an operator used for integrating deeply supervised layer features.

The loss function is defined in Equation~\ref{eq:type2_loss}. Total loss is the sum of final output layer loss and different deeply supervised loss. Similar to the aforementioned HLDS, a hyper-parameter $\alpha$ is used to control the balance between intermediate supervision loss and the final output layer loss.

\begin{align}
& \mathcal{L}=\mathcal{L}_L + \sum_{i=1}^{n}\alpha_{i} \times \mathcal{L}_i \label{eq:type2_loss}
\end{align}
where $n=3$ in Figure~\ref{fig:type2}, representing 3 branches.

In the backward propagation, for each branch, parameters are updated and learned by minimising the final output layer loss and that branch's deep supervision loss (Equation~\ref{eq:type2_back}).

\begin{align}
& \theta^{+}_{l}= \theta_{l}-\eta \times (\frac{\partial \mathcal{L}_{L}}{\partial \theta_{l}}+ \frac{\partial \mathcal{L}_{B}}{\partial \theta_{l}}) \label{eq:type2_back}
\end{align}


\noindent where $\theta_{l}$ is the parameter at the $l^{th}$ layer, $\eta$ is the learning rate, $\theta^{+}_{l}$ is the updated parameter at the $l^{th}$ layer, $\mathcal{L}_{L}$ is the loss at the $L^{th}$ layer (the final layer) and $\mathcal{L}_{B}$ is the loss at that branch's deep supervision loss.

The main idea behind DBDS is to make the final prediction by combining different branches' and levels' predictions. In DBDS, solving the gradient vanishing problem is not the main purpose, instead, exploring different level features is the main consideration. There is a consensus that different level features have different semantic meanings and contribute differently to the final task. In DBDS, the outputs from deep supervision at different branches and levels represent different predictions that different places of the neural network make, and the final prediction is made by integrating these intermediate predictions together.

\subsection{Category 3 - Deep Supervision Post Encoding (DSPE)}

In DSPE, the layer being deeply supervised is added back to the main neural network to encode extra information to the whole network.

\begin{figure}[ht]
\centering
\includegraphics[width=0.45\textwidth]{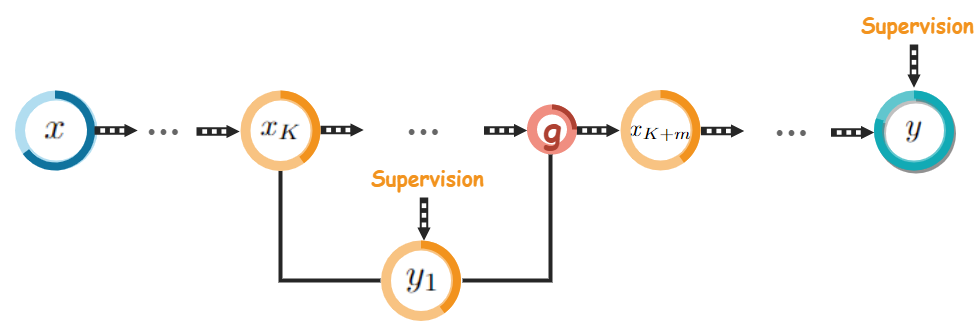}
\caption{Category 3 DSPE: the layer deeply supervised are encoded back to the main network.}
\label{fig:type3}
\end{figure}

In the forward propagation phase, as shown in Figure~\ref{fig:type3}, the deep supervision happens at $y_{1}$ after layer $K$, and after being supervised, it is encoded back to the main network by an operator $g()$. In this case, more discriminative features learned from deep supervision contribute back to the main network and affect the forward propagation. Equation~\ref{eq:type3_forward1} and~\ref{eq:type3_forward2} describe the forward propagation phase of Figure~\ref{fig:type3}.

\begin{align}
& y_{1}=f^{y_{1}}(x_{K}; \theta_{1}) \label{eq:type3_forward1} \\
& x_{K+m}=g(y_{1}, x_{K+m-1}) \label{eq:type3_forward2}
\end{align}
where $g()$ is a function to encode $y_{1}$ as part of inputs integrating into the following forward propagation. $g()$ can be concatenation~\cite{folle2019dilated,RN56}, elementwise-product~\cite{li2021parallel}, etc.

\begin{align}
& \mathcal{L}=\mathcal{L}_L + \sum_{i=1}^{n}\alpha_{i} \times \mathcal{L}_i \label{eq:type3_loss}
\end{align}
where $n=1$ in Figure~\ref{fig:type2}.

Similar to HLDS and DBDS, the loss function (Equation~\ref{eq:type3_loss}) here is also a combination of final layer loss and intermediate loss.

In backward propagation, parameters are updated and learned by minimising the final output layer loss and deep supervision loss.

\begin{align}
& \theta^{+}_{l}=\left\{
\begin{aligned}
\theta_{l}-\eta \times \frac{\partial \mathcal{L}_{L}}{\partial \theta_{l}} & , l=K+m,...,L.\\
\theta_{l}-\eta \times (\frac{\partial \mathcal{L}_{L}}{\partial \theta_{l}}+ \frac{\partial \mathcal{L}_{K}}{\partial \theta_{l}}) & , l=1,...,K+m-1.
\end{aligned}
\right. \label{eq:type3_back}
\end{align}
\noindent where $\theta_{l}$ is the parameter at the $l^{th}$ layer, $\eta$ is the learning rate, $\theta^{+}_{l}$ is the updated parameter at the $l^{th}$ layer, $\mathcal{L}_{L}$ is the loss at the $L^{th}$ layer and $\mathcal{L}_{K}$ is the loss at the $K^{th}$ layer.

The main idea behind DSPE is to encode the directly supervised features at hidden layers back to the main network to enhance the feature learning. The deeply supervised features can be used as attention maps~\cite{li2021parallel} to extract more discriminative features. 

\subsection{Summary}
All three categories keep the main advantages of deep supervision such as making the gradient flow more smoothly, meanwhile, their focuses are different. HLDS is the simplest method to employ the deep supervision, it aims to alleviate the gradient vanishing problem and also to provide some transparency for the network to some extent (deep supervision loss can be monitored in the process of training). HLDS does not explicitly use the deep supervision results as contributions to the final result. In contrast, DBDS explicitly takes advantage of the deep supervision results in different branches of the network to contribute to the final result. DBDS expects to fuse different branches' deep supervision results together to make the final output. Similar to DBDS, DSPE also explicitly uses the deep supervision results in the network. However, the differences between DBDS and DSPE are that in DBDS, deep supervision results are directly used for final result prediction, while in DSPE, deep supervision results are encoded back to the main network to continue forward propagation. In addition, the three categories of different deep supervision networks are not independent, they can be jointly used together for different purposes. We summarised the three categories of deep supervision works in the Applications section. 


\section{Learning Strategies}
There are different learning strategies when applying deep supervision. We discuss the learning strategies based on three aspects, i.e., the place to employ deep supervision, the combination of hidden layer loss and final layer loss, and the risk of deep supervision.

\subsection{The place to employ deep supervision}
Deep supervision can be embedded in different places of the neural network~\cite{newell2016stacked,lee2015deeply}. In general, the place where deep supervision is employed learns more discriminative features than other places because the features are more directly and closely supervised by the intermediate labels. Table~\ref{tab:places} lists different works applying deep supervision at different places of network. There are two advantages of adding deep supervision to different places of neural network.

The first advantage is that the final prediction is enhanced by including both shallow and deep layer outputs. Shallow layers learn low and local level features and deep layers learn high and global level features \cite{zeiler2014visualizing}. Therefore, combining shallow and deep layer results from deep supervision improves the final prediction performance. Lee et al. \cite{lee2015deeply} employed deep supervision at shallow layers of the neural network to do the image classification, and the loss from deep supervision was combined together with the final prediction layer loss to the main objective function. This is to obtain explicit classification output from different levels of the network. Wang et al. \cite{wang2015training} added auxiliary supervision branches after certain intermediate layers during training and formulated a simple rule to determine where these branches should be added to improve the classification performance. In addition, Newell et al. \cite{newell2016stacked} generated a stack hourglass network for keypoint detection problems. In this work, deep supervision was added at the end of each stack, representing different levels of information, while the loss is combined in the main objective function. Table \ref{tab:places} summarises the works in which deep supervision is added at different places of the neural network. 

\begin{table*}[ht]
\centering
{\scriptsize
\caption{Summary of adding deep supervision into different places of the network.}
\begin{tabular}{p{0.8in}|p{1.5in}|p{1.5in}|p{1.5in}|p{0.8in}} \hline 
Places & Description & Advantages & Disadvantages & Typical examples \\ \hline \hline
Shallow layers & Integrated direct supervision to the earlier layers  & Help to assist low-level feature learning and make it consistent with high-level feature learning\newline \newline  & The features of lower layers are generic, and the representation power of low layers is weak. This might pose a risk of insufficient representation learning. & \cite{lee2015deeply}\cite{dou20173d} \\ \hline 
Intermediate layers & Add supervision to intermediate network layers and initialise the lower convolution layers and the rest of the layers. & Reduce training and verification errors and granting the network better convergence\newline & Lack coarse-level features from early layers which are beneficial for image-level classification, while directly adding simple auxiliary classifiers to the intermediate layers might not be beneficial. & \cite{wang2015training} \cite{chen2016deep} \cite{liu2020encoder} \\ \hline 
Deep layers & Apply deep supervision to the bottom layers. & The bottom sides of the neural network can learn discriminate features and help find fine details, thus making the final results smoother.  & May cause distinct supervision targets. & \cite{newell2016stacked} \cite{liu2022semantic} \\ \hline 
\end{tabular}
\label{tab:places}
}
\end{table*}

The second advantage is that deep supervision provides more transparency through learning. The loss at each deep supervision layer can be used as an indicator of how gradients are changing during the learning process, which can further help select the optimal place to embed the deep supervision. During our search of the literature, we did not find any papers studying the optimal place where deep supervision should be employed. However, we believe that the places where deep supervision is embedded can be dynamically changed and the optimal place of deep supervision can be determined by monitoring the loss changes and gradient changes during the learning process.

\subsection{The combination of hidden layer loss and final layer loss}
When applying deep supervision in the neural network, it will bring different deep supervision-related loss to the main objective function. In general, a hyper-parameter $\alpha_{i}$ is used to control the balance between the final layer's loss and hidden layers' loss \cite{lei2020lightweight}. There are different strategies to set $\alpha_{i}$ (Table~\ref{tab:balance}). 

\begin{align}
& \mathcal{L}=\sum_{i=1}^{n}\alpha_{i} \times \mathcal{L}_i \label{eq:alpha_loss}
\end{align}
where $\mathcal{L}_i$ represents different deep supervision related loss.

$\alpha_{i}$ in Equation~\ref{eq:alpha_loss} can reflect the weights of different loss included in the main objective function. The higher the weight for a specific loss term, the more attention that the objective function will pay to in the learning process. For example, Li et al. \cite{li2021parallel} set $\alpha=1$ in the objective function, indicating that in the objective function, multiple loss is evenly allocated. In this case, deep supervision and final layer supervision are considered to play equal roles in the learning process. $\alpha_{i}$ can also be set by a sequence of weight values, indicating that the fixed percentages of multiple loss, such as in \cite{lei2020lightweight}, variant balancing weights were used for four loss by using fixed distribution as 10\%, 20\%, 30\% and 40\% respectively. Wu et al. \cite{wu2021attention} designed an attention deep model with multi-scale deep supervision for person re-identification. Deep supervision is employed at different scales and there is five deep supervision loss with weights being set to be 0.4, 0.1, 1, 0.03 and 0.03. In most cases, deep supervision loss is allocated lower weights than the final layer loss, as the deep supervision loss is regarded as auxiliary loss in the main objective function, and the final layer loss plays a dominant role in the optimising process~\cite{wu2021attention}. The selection of $\alpha_{i}$ is not limited to just these cases. $\alpha_{i}$ can be dynamically adjusted during the learning process based on the performance of the validation dataset as in \cite{hou2017deeply} and \cite{dou20173d}, which provides a more flexible learning strategy.

We conclude that there are three principles that can be used to determine the value of $\alpha_{i}$, shown in Table \ref{tab:balance}. The first principle is to evenly set the $\alpha_{i}$ in the learning process. The second principle is to allocate weights by non-equal balancing with different scales. The third principle is to allocate weights by monitoring the loss change during the learning process. For example, if some loss is changing very slightly in the learning process, to reflect such changes, more weights can be allocated to these terms, and vice versa. Dynamic adjusted $\alpha_{i}$ is an appropriate option in most cases because it provides a flexible option. 

\begin{table*}[ht]
\centering
{\scriptsize
\caption{Summary of different strategies to set balance parameters.}
\begin{tabular}{p{1.4in}|p{2in}|p{2in}} \hline 
Learning strategy & Typical explanation & References \\ \hline \hline
Equal balancing  & The~parameters~are~fixed~values during the training procedure & [25][28][31][33][38][47][68][72][77][79][84] [88][90][104][109][111][114][116][118][122][128]\newline[131][133][134] \\ \hline 
Fixed-form distribution & Uniformed distribution when combining classes & [42][51][54][56][60][83][106][107][129] \\ \hline 
Dynamic adjustment & Decaying by a specific percentage for every specific epoch during learning. & [18][40][51][52][63][64][89][100][105] \\ \hline 
\end{tabular}
\label{tab:balance}
}
\end{table*}

Apart from different combination of loss, there are also different weights updating strategies in the learning process. On one hand, hidden layer loss and the final layer loss can be updated simultaneously during learning \cite{mai2019human}. Mai et al. \cite{mai2019human} used ResNet50 as backbone to form a three-stage deep supervision network to estimate human pose, while loss was added together by setting $\alpha_{i}=1$ in the objective function and updated simultaneously. This strategy is adopted when all loss continues to decrease during the learning process epoch by epoch. On the other hand, some shallow hidden layer loss is updated in the first couple of epochs and frozen afterwards, meanwhile, deep hidden layers and the final layer loss are updated over all epochs. We haven't found any work use this strategy, but we believe that this strategy can take advantage of the `transparency' given by deep supervision to avoid the overfitting of shallow hidden layers' weights learning. The weights updating strategies are quite flexible and it is hard to identify an optimal strategy. The neural network performance can be improved by fine-tuning different hyper-parameters and weights updating strategies.

\subsection{The risk of deep supervision in learning}


Although deep supervision improves neural network performance, it may lead to some risks. Very few papers have mentioned the risks of deep supervision. In this review paper, we outline some risks of deep supervision.

The first risk is the overfitting problem. When applying deep supervision, since the weights are updated through backward propagation by minimising all loss, this may lead to the update of some weights being overwhelmingly supervised by minimising the multiple loss (including the final layer loss and deep supervision-based loss). In this case, such learned weights tend to make the whole network to be overfitting.

The second risk is the computation cost. When introducing more loss to the learning process, it is likely to significantly increase the training time and inference time of the network by calculating more loss and gradients in the learning process.

The third risk is the issue of tuning hyper-parameters. When introducing deep supervision into a neural network, it needs more hyper-parameters. The way to select and tune such hyper-parameters is based on a large number of experiments and it is task-specific. In general, it is hard to set a common rule for the selection of such hyper-parameters. For instance, it is hard to set hyper-parameter $\alpha_{i}$ values for controlling the balance between the final loss and deep supervision loss. Different $alpha_{i}$ values may have a different impact on the training results. Even if we decide to give more weights on the final layer loss and fewer weights on the intermediate layer loss, how `more' and how `less' still needs a large number of experiments to prove.

\section{Applications}
In this section, we reviewed different deep supervision applications in various areas, summarised it in Table \ref{tab:application-total}.

\begin{table*}[!ht]
\centering
{\scriptsize
\caption{Applications of deep supervision.}
\begin{tabular}{p{0.8in}|p{2in}|p{3in}} \hline 
Application field & Task & Typical Datasets \\ \hline \hline 
Image segmentation & Semantic segmentation\newline Instance segmentation & CT scan images \cite{zhou2017deep}\newline MRI images \cite{lin2021fully}\newline Ultrasound images\cite{mishra2018ultrasound} \\ \hline 
Key point detection & human pose estimation \newline face keypoint detection \newline hand keypoint detection & COCO dataset and MPII Human Pose dataset \cite{mai2019human}\newline AFLW-PIFA for large pose face alignment \cite{bulat2016convolutional} \newline CMU Panoptic Dataset, Onehand10K Dataset and HGR1 Dataset \cite{folle2019dilated} \\ \hline 
Image classification & Pixel-based image classification\newline Patch based image classification\newline Feature-based image classification & CIFAR100 \cite{ai2017human} \newline SVHN database \cite{lee2015deeply}\newline HEp-2 cell images \cite{lei2018deeply}  \\ \hline 
Object detection & Object detection in images\newline Moving object detection in videos\newline Salient Object Detection & INRIA pedestrian database \cite{al2016novel}\newline PASCAL VOC 2007\cite{zou2019object}\newline DUTO, HKU-IS, ECSSD, PASCAL-S, SOD and DUTS \cite{lei2020lightweight}\cite{fan2021u}  \\ \hline 
Super-resolution & SR algorithms in the spatial domain SR algorithms in the frequency domain & DIV2K dataset \cite{lai2018fast}\newline Set5, Set14, B100 and Urban100 \cite{kim2016deeply} \\ \hline 
\end{tabular}
\label{tab:application-total}
}
\end{table*}

\subsection{Image Segmentation}
\begin{figure*}[ht]
\centering
\includegraphics[width=0.6\textwidth]{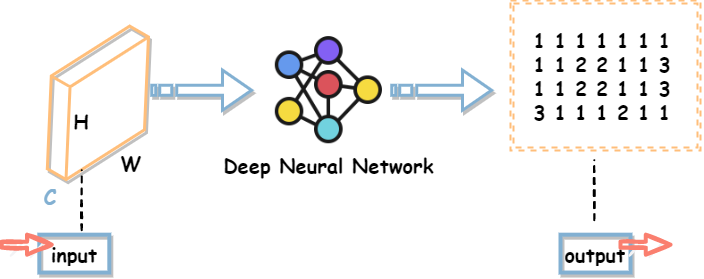}
\caption{Image segmentation task, with the input image as an $H \times W \times C$ tensor ($C = 3$, representing an R,G,B image) and the output $H \times W \times 1$ tensor where each pixel value represents the class it belongs to.}
\label{fig:is}
\end{figure*}

Image segmentation, describing the process of partitioning images into multiple segments, is a key task in computer vision. Semantic segmentation and instance segmentation, as two main categories of image segmentation, detects object category and individual object respectively from an image. In image segmentation, the input can be an RGB image with a shape of $H \times W \times 3$ or a grayscale image with a shape of $H \times W \times 1$, and the output is a segmentation map (with a shape of $H \times W \times 1$) where each pixel value represents the class it belongs to. The label shape is also an $H \times W \times 1$ tensor. Figure~\ref{fig:is} shows the image segmentation task and Table~\ref{tab:image-segmentation} shows different works of image segmentation using deep supervision.

There is an emerging use of image segmentation with deep supervision on medical images and videos, such as CT scans, MRI and ultrasound - commonly used techniques assisting clinicians with diagnosis and tracking the progress of diseases. Mishra et al. \cite{8502126} proposed a deeply supervised network for ultrasound image segmentation. The proposed network used a backbone similar to VGG-16 and was deeply supervised along with the convolutional layers of the backbone. In detail, the auxiliary layers branched from the backbone were firstly supervised by the label, and then fused to generate boundaries (i.e., output) of the input images. The designed structure of deep supervision forced the model to learn both deep features and shallow features with the assistance of labels. The labels used for deep supervision and final layer supervision were the same (category 1 HLDS). The proposed deeply supervised network achieved 0.8 mIoU. Similarly, deep supervision applications on ultrasound images also demonstrated an improvement in evaluation metrics of the model \cite{wang2019deeply}. The model proposed by Wang et al. \cite{wang2019deeply} adopted a U-Net backbone, with a densely deep supervision (DDS) mechanism forcing the model to learn discriminative characteristics of the target object (i.e., breast cancer). The structure of U-Net allowed the DDS to be applied to multi-scale features, which achieved 0.95$\pm$0.20 sensitivity of cancer detection from ultrasound. This work followed category 1 - HLDS and made the gradient propagate more smoothly to improve the transparency of the network. Mo et al. \cite{mo2017multi} employed an encoder-like network structure to segment the retinal vessel. The network learned different scale feature maps (from the scale of $512 \times 512$ to the scale of $64 \times 64$), and deep supervision is employed at different levels and branches of the main network. The pixel-level probability map was used to supervise intermediate and final outputs. The network achieved a sensitivity of 0.7661$\pm$0.0533 by fusing intermediate supervision outputs and final layer outputs together. This is a common practice of using deep supervision to segment medical images by following category 2 - DBDS. Depending on the intended shape of intermediate outputs, different scale probability heatmaps can be used as labels in deep supervision. In addition, category 3 - DSPE also has been extensively adopted in medical image segmentation. Nie et al. \cite{nie2020spatial} proposed a Spatial Attention-based Efficiently Feature Fusion Network (SA2FNet) to segment brain tumors based on an MRI dataset. SA2FNet adopted an encoder-decoder-like structure and refined low-level spatial features in the proposed spatial attention module. Deep supervision was employed in the spatial attention module to make the network learn a more discriminative attention map, and the learned attention map was encoded back to the main network to guide the main network to focus more on the regions of segmentation (Category 3 - DSPE). SA2FNet achieved a segmentation dice score of 90.62 on Brain Tumor Segmentation (BraTS) 2018 challenge dataset. The application of deep supervision is not limited to medical images, and it has been used in coastline detection in geoscience. Heidler et al. \cite{RN128} proposed a HED-UNet to segment the coastline based on the images taken by optical and synthetic radar sensors from the Antarctic coastline. HED-UNet adopted an encoder-decoder structure learning pyramid features for the segmentation. Deep supervision was applied in the decoder part, aiming to explicitly encode meaningful features in the deep layers of the main network. HED-UNet achieved 92.0\%$\pm$0.8\% accuracy on the Wilkes Land dataset and 80.5\%$\pm$1.6\% accuracy on the Antarctic Peninsula dataset.

Three categories of deep supervision discussed above have their application in image segmentation. Ground truth labels, in image segmentation applications, can be easily used as deep supervision labels as their shapes are exactly the same. Generally, category 1 HLDS, category 2 DBDS, and category 3 DSPE, use labels with the shape of $H \times W \times 3$ (RGB image) or the shape of $H \times W \times 1$ (greyscale image). Depending on the backbone structure used, deep supervision supervises multi-scale features or shallow features and deep features, and improves sensitivity to image segmentation. 

\begin{table}[!ht]
\centering
{\scriptsize
\caption{Deep supervision embedded image segmentation works.}
\begin{tabular}{p{0.35in}|p{0.6in}|p{0.75in}|p{0.9in}}
\hline
Category & Label & Loss Sharing & Works \\
\hline
\hline
\multirow{8}{*}{HLDS} & \multirow{4}{*}{Same labels} & Pro rata & \cite{dou20163d,asl2018alzheimer,guo2019bts,8374823,8759535,wang2019deeply,8802952,guo2020image,xia2020extracting,mathur2020deep,lei2019ultrasound}  \\
& & Dynamic pro rata & \cite{8733792,zhou2019deep} \\
& & Evenly \\
& & No disclosure & \cite{dolz2018isointense,liu2020encoder} \\
& \multirow{4}{*}{Different labels} & Pro rata & \cite{RN47}  \\
& & Dynamic pro rata & \cite{RN76,8965924} \\
& & Evenly \\
& & No disclosure \\
\hline
\multirow{8}{*}{DBDS} & \multirow{4}{*}{Same labels} & Pro rata & \cite{lei2020lightweight,zeng20173d,chatterjee2020ds6,bortsova2017segmentation,8502126,samuel2019multilevel,yang2020dual,wang20203d} \\
& & Dynamic pro rata & \cite{ma2019iterative,yang2018combating} \\
& & Evenly & \cite{DONG2019192} \\
& & No disclosure & \cite{chen2020fully,li2019p3,qamar2019multi} \\
& \multirow{4}{*}{Different labels} & Pro rata & \cite{mo2017multi,yang2017towards,zhou20203d,wang2019deeply,li2020unified,li2020deeply,habijan2020abdominal,zhao2020mss,zhou2017deep,zhao2019multi}  \\
& & Dynamic pro rata & \\
& & Evenly & \\
& & No disclosure \\
\hline
\multirow{8}{*}{DSPE} & \multirow{4}{*}{Same labels} & Pro rata &  \\
& & Dynamic pro rata & \cite{folle2019dilated,8946023,kearney2019attention,RN85} \\
& & Evenly &  \\
& & No disclosure & \cite{RN56} \\
& \multirow{4}{*}{Different labels} & Pro rata & \cite{nie2020spatial,mo2020iterative,zhang2020multi,RN19} \\
& & Dynamic pro rata & \cite{RN43,RN41}  \\
& & Evenly & \cite{RN161} \\
& & No disclosure & \cite{RN66,RN57,RN76} \\
\hline
\end{tabular}
\label{tab:image-segmentation}
}
\end{table}

\subsection{Keypoint Detection}
\begin{figure*}[ht]
\centering
\includegraphics[width=0.8\textwidth]{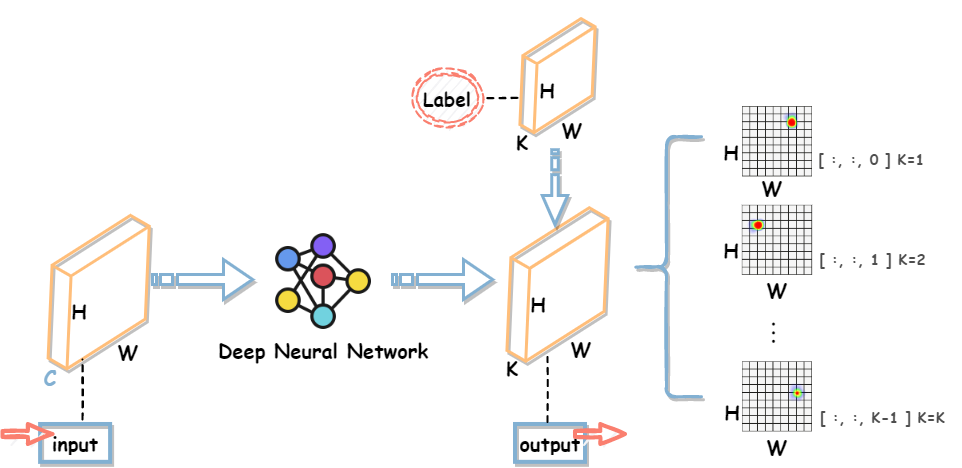}
\caption{Keypoint detection task, with the input image as an $H \times W \times C$ tensor ($C = 3$, representing an R,G,B image) and the output heatmap as an $H \times W \times K$ tensor.}
\label{fig:kd}
\end{figure*}

Keypoint detection from images is a challenging research topic in computer vision that aims to identify the locations of important object parts. There are a wide range of applications for keypoint detection, including human pose estimation \cite{newell2016stacked,mai2019human}, face keypoint detection \cite{bulat2016convolutional}, and hand keypoint detection \cite{li2021parallel}. In general, keypoint detection is regarded as a prediction task to map an image input to heatmap-based keypoint positions in the image. In particular, the input image shape is $H \times W \times C$, the output heatmap shape is $H \times W \times K$ and labels used to supervise the neural network are $2D$ Gaussian response heatmaps with the shape of $H \times W \times K$, where $C = 3$ in most cases representing the $R, G, B$ 3 channels and $K$ refers to the number of types of the keypoints, $H, W$ represents the input image's or label's (Gaussian response heatmap) height and width. Figure~\ref{fig:kd} shows the process of keypoint detection and Table~\ref{tab:keypoint-detection} lists the different works of keypoint detection using deep supervision.

Many works apply different categories of deep supervision to CNNs for improving keypoint detection performance. Newell et al. \cite{newell2016stacked} designed a stacked hourglass-shaped CNN to estimate human pose, which consisted of many encoder-decoder (hourglass-shaped) structures that were connected together. Deep supervision was applied between the two consecutive stacks of hourglass-shaped structures of their proposed network. In particular, the purposes of deep supervision were twofold. The first was to smooth the gradient flow and the second was to encode the results being deeply supervised back to the network (between two consecutive stacks of hourglass-shaped structures). In this way, different stacks of the hourglass-shaped structure refined the prediction by integrating each stack's deep supervision results together through the learning, and the final layer output was used for keypoint prediction. Their approach achieved state-of-the-art performance in human pose estimation on the MPII Human Pose dataset, with elbow accuracy being 91.2\% and wrist accuracy being 87.1\%. Stacked hourglass-shaped CNN is a successful application of deep supervision of category 3 - DSPE. However, there are different ways to encode deep supervision results to the main network in category 3 - DSPE. Li et al. \cite{li2021parallel} designed a parallel multi-scale deep supervision network to detect hand keypoints. In their proposed neural network, deep supervision was applied at different branches, and the results from deep supervision were encoded back to the main network. The purpose of the deep supervision was to help create a spatial attention map, which was encoded back to the main network by applying element-wise multiplication between the feature map in the main network and the spatial attention map created by the deep supervision. In this way, the feature map in the main network was activated by enhancing the keypoint surrounding area features and weakening the non-keypoint area features with the help of the spatial attention map. This approach was effective in hand keypoint detection, achieving the state-of-the-art performance on OneHand10K, CMU and HGR1 hand datasets. The aforementioned works apply category 3 - DSPE deep supervision paradigm, however, category 2 - DBDS is also commonly used in keypoint detection problems. In the studies of \cite{8965911} and \cite{8633110}, deep supervision was employed at different branches of the neural network and the final human pose prediction is made by integrating the deep supervision results of different branches. The purpose of deep supervision was to directly provide prediction at different places of the neural network. In this way, different results from deep supervision were integrated to do the final prediction. This approach also proved to be effective in human pose estimation and achieved state-of-the-art results in MPII and COCO datasets. Table~\ref{tab:keypoint-detection} lists different deep supervision embedded keypoint detection works.

CNNs designed for solving keypoint detection problems are either very deep (with different stacks) or use multi-scale or multi-branch networks, where deep supervision can take effect by not only making the gradient spread more smoothly but also embedding different levels' supervision results or different scale supervision results into the main network, which improves the performance of keypoint detection. Additionally, the intermediate feature maps in the CNNs have similar shapes ($h \times w \times s$) (where $h$, $w$ are the height and width of the feature map and $s$ is the number of kernels) with the labels ($H \times W \times K$) (where $H$ and $W$ are height and width of the scaled image and $K$ is the number of keypoints) used in deep supervision, so it is easy for the results from the deep supervision being embedded back (through scaling) to the CNNs (category 3 - DSPE) or being combined with other layers' feature maps together for the final prediction (category 2 - DBDS).

\begin{table}[!ht]
\centering
{\scriptsize
\caption{Deep supervision embedded keypoint detection works.}
\begin{tabular}{p{0.35in}|p{0.6in}|p{0.75in}|p{0.9in}}
\hline
Category & Label & Loss Sharing & Works \\
\hline
\hline
\multirow{8}{*}{HLDS} & \multirow{4}{*}{Same labels} & Pro rata &  ~\cite{wu2019lightweight} \\
& & Dynamic pro rata \\
& & Evenly \\
& & No disclosure & ~\cite{liu2020robust} \\
& \multirow{4}{*}{Different labels} & Pro rata &  \\
& & Dynamic pro rata \\
& & Evenly \\
& & No disclosure \\
\hline
\multirow{8}{*}{DBDS} & \multirow{4}{*}{Same labels} & Pro rata &  ~\cite{qian2020cephann,8633110,8434117,9033204}\\
& & Dynamic pro rata \\
& & Evenly \\
& & No disclosure \\
& \multirow{4}{*}{Different labels} & Pro rata & ~\cite{li2017deep,ke2018multi}\\
& & Dynamic pro rata &  \\
& & Evenly & ~\cite{8965911}\\
& & No disclosure \\
\hline
\multirow{8}{*}{DSPE} & \multirow{4}{*}{Same labels} & Pro rata & ~\cite{ai2017human} \\
& & Dynamic pro rata \\
& & Evenly & ~\cite{newell2016stacked} \\
& & No disclosure & ~\cite{wang2020landmarknet,bulat2016convolutional} \\
& \multirow{4}{*}{Different labels} & Pro rata & ~\cite{RN18} \\
& & Dynamic pro rata \\
& & Evenly \\
& & No disclosure \\
\hline
\end{tabular}
\label{tab:keypoint-detection}
}
\end{table}

\subsection{Image Classification}
Image classification is a computer vision task aiming to classify different images into correct classes. Particular classes can be `airplane', `bird', etc \cite{cifar10}. There are a wide range of applications from image classification, including cell classification \cite{lei2018deeply}, face recognition \cite{liu2018learning}. In general, image classification can be regarded as a classification problem to map an image input to a vector that represents the probabilities of each class the image belongs to. And the image is predicted as a particular class with the highest probability in the vector. In particular, the input image shape is $H \times W \times C$, the output is a $N \times 1$ vector activated by the SoftMax function (mapping each element in the vector to a probability) and the label used to supervise the neural network also has the shape of $N \times 1$, where $C=3$ in most cases representing the $R, G, B$ 3 channels and $N$ refers to the number of classes, $H, W$ represent the input image height and width.

\begin{figure*}[ht]
\centering
\includegraphics[width=0.5\textwidth]{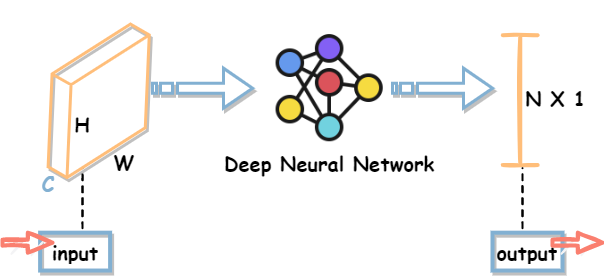}
\caption{Image classification task, with the input image as an $H \times W \times C$ tensor ($C = 3$, representing an R,G,B image) and the output as an $N \times 1$ tensor ($N$ represents the number of class).}
\label{fig:ic}
\end{figure*}

Deep supervision has been applied to different backbone networks for image classification. Lee et al. \cite{lee2015deeply} designed a deeply supervised neural network to classify the image, in which early deep supervision was applied on different hidden layers and the final prediction was made by the final layer output but constrained by the discriminative features deeply supervised at those hidden layers during the learning process. The purpose of deep supervision was to first let some hidden layers learn more discriminative features, so that the classifier trained on these highly discriminative features, and secondly alleviate the gradient vanishing problem. The experiment also showed that the direct pursuit of feature discriminativeness led by deep supervision would not affect the overall network performance. Deep supervision was proved to be effective in their study and achieved a 1.92\% error rate on Street View House Numbers (SVHN) database. Lei et al. \cite{lei2018deeply} applied deep supervision on the hidden layers of their proposed ReNet50 network to classify the HEp-2 cell images. Similar to \cite{lee2015deeply}, the purpose of deep supervision was to train discriminative features in hidden layers expecting to use such features to improve the classification performance. In particular, deep supervision was employed at low and mid-level of hidden layers. The outputs at these hidden layers were not integrated together to do the final prediction, instead, only the final layer output was used as the prediction (which already considered the information learned at those hidden layers where deep supervision was applied). This approach also proved to be effective in cell image classification and achieved 97.14\% accuracy on ICPR2012 dataset. Table \ref{tab:image-classification} lists different image classification works using deep supervision.

Category 1 - HLDS and category 2 - DBDS are commonly used in solving image classification tasks. Generally, both HLDS and DBDS deep supervision structures are beneficial in two aspects. First, deep supervision improves the smoothness of gradient backward propagation, so that to some extent, it alleviates the gradient vanishing issue. Second, the shallow level deeply supervised features provide more superior determinativeness to deep layers (both HLDS and DBDS). Category 3 - DSPE is barely used because of the different shapes between label and feature maps. In image classification tasks, the label used is a vector with the shape of $N \times 1$, while the feature map in the neural network is a tensor with the shape of $H \times W \times C$ (where $H$, $W$ are height and width of the feature map and $C$ is the number of kernels). In the structure of category 3 - DSPE, it is hard to encode the deeply supervised output (with the shape of $N \times 1$) back to the main network whose feature map is in the shape of $H \times W \times C$. 

\begin{table}[!ht]
\centering
{\scriptsize
\caption{Deep supervision embedded image classification works.}
\begin{tabular}{c|l|l|l}
\hline
Category & Label & Loss Sharing & Works \\
\hline
\hline
\multirow{8}{*}{HLDS} & \multirow{4}{*}{Same labels} & Pro rata &   \\
& & Dynamic pro rata \\
& & Evenly & \cite{lee2015deeply} \\
& & No disclosure &  \\
& \multirow{4}{*}{Different labels} & Pro rata &  \\
& & Dynamic pro rata \\
& & Evenly \\
& & No disclosure & \cite{RN137} \\
\hline
\multirow{8}{*}{DBDS} & \multirow{4}{*}{Same labels} & Pro rata & \cite{al2016novel,lei2018deeply}  \\
& & Dynamic pro rata \\
& & Evenly \\
& & No disclosure \\
& \multirow{4}{*}{Different labels} & Pro rata & \cite{liu2018learning} \\
& & Dynamic pro rata & \\
& & Evenly & \\
& & No disclosure \\
\hline
\end{tabular}
\label{tab:image-classification}
}
\end{table}

\subsection{Object Detection}

Object detection is one of the most important computer vision tasks, which involves detecting instances of visual objects and locating their positions in an image. It is also the basis of other important computer vision tasks, including instance segmentation, image captioning, object tracking, etc \cite{zou2019object}. Object detection techniques are widely used in a number of real-world scenarios, such as pedestrian detection, face detection, hand-writing letters detection, remote sensing target detection, traffic sign or lights detection, etc \cite{zou2019object}. Salient object detection is slightly different from object detection and it can be described as a combination of object detection and instance segmentation. Salient object detection can be further divided into 1) detecting the most salient object in a visual image and 2) segmenting the identified object from the image. Table \ref{tab:object-detection} lists different object detection tasks using deep supervision. Deep supervision has shown effectiveness in both traditional object detection and salient object detection works.

\begin{figure*}[ht]
\centering
\includegraphics[width=1\textwidth]{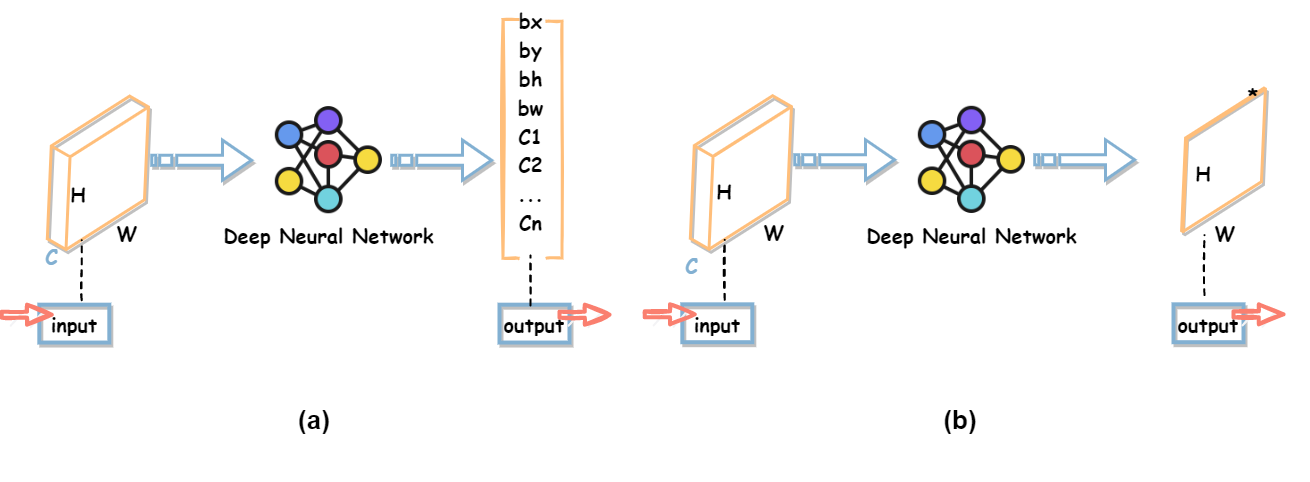}
\caption{Object detection task (a). with the input image as an $H \times W \times C$ tensor ($C = 3$, representing an R,G,B image), the output will include $b_x$,$b_y$ represent the coordinates of the bounding boxes, $b_w$ and $b_h$ represent the width and height of the bounding boxes, $C_n$ represents the prediction of $n$ number of classes. Salient object detection (b). with the input image as an $H \times W \times C$ tensor, the output image only contain pixel 0 or 1, representing the probability of the pixel belonging to the detected object.}
\label{fig:od}
\end{figure*}

For traditional object detection tasks, the input of a model is a visual image with the shape $H \times W \times 3$ ($H$ and $W$ as the shape of the input image and 3 representing R, G, B channels) and the output is usually an image consisting of the following information: the coordinates of bounding boxes (`where is the object?'), the class of the object (`what is the object?'), and the probability score (`how confident is the prediction?'). Li et al. \cite{li2019real} proposed a multi-scale based neural network using deep supervision to achieve real-time pedestrian detection. In their work, deep supervision was applied at each prediction layer with different scales to guide the side outputs toward region predictions with the desired characteristics. This type of deep supervision has shown improvement in optimisation and generalisation for object detection tasks. This approach is a successful application of deep supervision category 1 - HLDS, which achieved state-of-the-art performance on the INRIA pedestrian dataset. It indicates the application of deep supervision works well with multi-scale based neural networks for object detection tasks. Shen et al. adopted deep supervision in two of their studies \cite{shen2017dsod,shen2019object} by adding side-output layers to introduce an `auxiliary' objective at each hidden layer for the detection task, where they used classification loss and localisation loss for supervision. These works belong to category 2 - DBDS, and the same labels are commonly used for deep supervision. Based on their ablation studies, deep supervision was proved to increase the performance of object detection. Results showed that the performance increased on PASCAL VOC 2007 object detection dataset with the adoption of deep supervision. Hao et al. \cite{hao2020labelenc} also used the auxiliary loss to encourage the latent features generated by the backbone network for some `ideal' embedding, which is an example of category 3 - DSPE with different labels. The main idea of this method in object detection tasks is to add direct supervision to the earlier layers rather than the final output so that the gradient vanishing problem can be mitigated through some `auxiliary' objective functions.  

For salient object detection tasks, the input of a model is a visual image with shape $H \times W \times 3$ and the output is an image of the same size, with each pixel of the input image having a value in the range from 0 to 1, representing the probability that the pixel belongs to the salient object. Deep supervision is also widely used for solving salient object detection problems \cite{hou2017deeply,zhang2019salient}. Zhang et al. \cite{zhang2019salient} adopted a method called multi-layer intermediate supervision in the Resnet-50 network. In Resnet-50, the deeper layers usually include more semantic and global information whereas the skip connections in Resnet often merge those important spatial and semantic information. In this work, intermediate supervision was an elegant solution to ensure that contextual and global information can be accurately captured through the multi-layer intermediate supervision method. To achieve this, Zhang et al. \cite{zhang2019salient} used a cross-entropy loss to supervise the side-output in different scales of the feature map. This type of salient object detection study usually adopts category -2 DBDS with the same labels used for deep supervision. Compared to traditional object detection tasks, salient object detection is more difficult, which requires DNNs to capably extract distinctive objects or regions from the input images. 

\begin{table}[!ht]
\centering
{\scriptsize
\caption{Deep supervision embedded object detection works.}
\begin{tabular}{c|l|l|l}
\hline
Category & Label & Loss Sharing & Works \\
\hline
\hline
\multirow{8}{*}{HLDS} & \multirow{4}{*}{Same labels} & Pro rata & \cite{li2019real,RN48,li2020efficient,li2018object} \\
& & Dynamic pro rata \\
& & Evenly \\
& & No disclosure &  \\
& \multirow{4}{*}{Different labels} & Pro rata & \cite{le2017deeply}   \\
& & Dynamic pro rata \\
& & Evenly \\
& & No disclosure \\
\hline
\multirow{8}{*}{DBDS} & \multirow{4}{*}{Same labels} & Pro rata & \cite{shen2017dsod,zhang2019salient,shen2019object} \\
& & Dynamic pro rata & \cite{hou2017deeply,RN168} \\
& & Evenly & \cite{RN55} \\
& & No disclosure \\
& \multirow{4}{*}{Different labels} & Pro rata & \cite{RN109} \\
& & Dynamic pro rata &  \\
& & Evenly & \\
& & No disclosure & \cite{RN44} \\
\hline
\multirow{8}{*}{DSPE} & \multirow{4}{*}{Same labels} & Pro rata &  \\
& & Dynamic pro rata \\
& & Evenly & \\
& & No disclosure & \\
& \multirow{4}{*}{Different labels} & Pro rata & \cite{hao2020labelenc} \\
& & Dynamic pro rata \\
& & Evenly \\
& & No disclosure \\
\hline
\end{tabular}
\label{tab:object-detection}
}
\end{table}

\subsection{Super-Resolution}

\begin{figure*}[ht]
\centering
\includegraphics[width=0.65\textwidth]{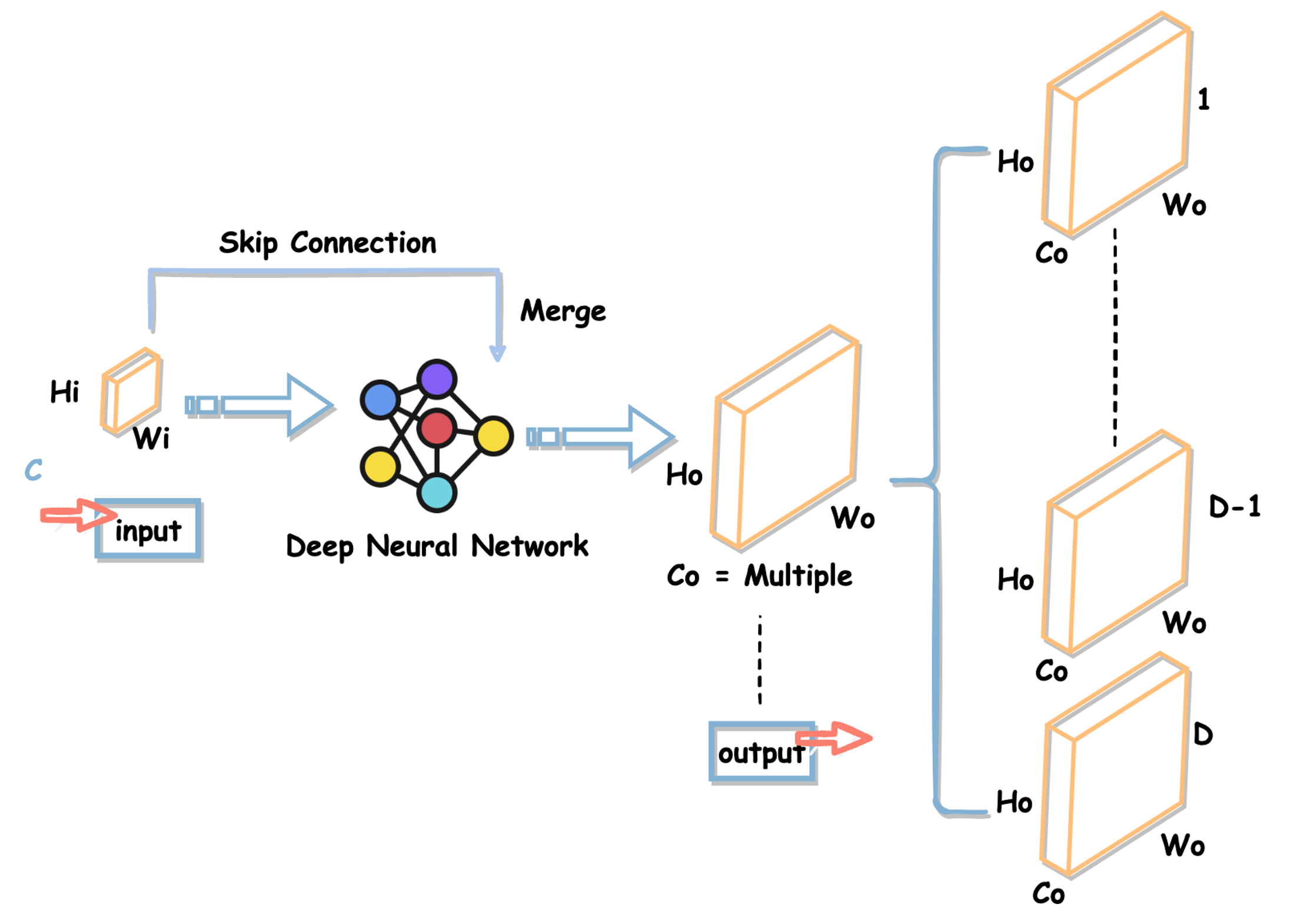}
\caption{Super-resolution task, with the input image as an $H \times W \times C$ tensor ($C = 1, 3$, representing a gray-scale or an R,G,B image) and the output as an $H \times W \times C$ tensor.}
\label{fig:sr}
\end{figure*}

Super-resolution (SR) is an inverse problem that aims to reconstruct high-resolution (HR) images from the corresponding low-resolution (LR) counterparts with image quality degradations. Super-resolution task is involved in various critical applications in different domains, such as face recognition in surveillance footage \cite{mudunuri2015low}, small object detection in scenes \cite{girshick2015region}, medical imaging \cite{oktay2017anatomically}, forensics \cite{ghazali2012super}, and remote sensing \cite{lillesand2015remote}. For SR as one of the image restoration problems, image details are critical, therefore the approaches that widen the receptive field such as using a pooling layer to reduce the intermediate representation dimension are utilised. Generally, SR can be considered as a prediction task of reconstructing a low-resolution (LR) input image based on feature maps or depth maps to deliver a high-resolution image prediction as the final output. Specifically, the input LR image shape is $H \times W \times C$, the output HR image shape is $H \times W \times C$ and the relatively soft labels used to supervise the neural network are $D$ predictions, where $C = 1 \text{ or } 3$ representing grayscale channel only or $R, G, B$ channels for input LR image and multiple channels for output SR image, and $H, W$ represents the height and width. Figure~\ref{fig:sr} illustrate the process of SR in the most recent relevant studies.

Applied to tackling the image SR topic in recent years, most studies are conducted from the DBDS approach. Kim et al. \cite{kim2016deeply} proposed a deeply recursive convolutional network (DRCN) with up to 16 recursions deep recursive layer that improved performance for image super-resolution. This DRCN applies the same supervised convolutional layer with the same reconstruction, avoiding the conventional over-fitting and over-size problems caused by adding a new weight layer to increase depth. Their model adopted a DBDS approach that supervises all recursions to alleviate the vanishing/exploding gradients effect, and during training, all predictions are simultaneously supervised, while the final output from the ensemble is supervised as well. The proposed advanced model in \cite{kim2016deeply} outperformed five existing models on four datasets, including two benchmark datasets Set5 \cite{bevilacqua2012low} and Set14 \cite{zeyde2010single}, with the average SSIMs from 0.7233 to 0.9588 for scale factor $\times 2,\times 3$ and $\times 4$. Most SRCNN simply stack convolutional weight layers as many times as possible to make the network deeper and wider, which significantly increases the number of parameters for training, making the model becoming too huge to be stored and retrieved, meanwhile requiring more data to prevent over-fitting. Instead of stacking convolutional layers to adopt a deep supervision scheme for SR, Huang et al. \cite{8853318} introduced a pyramid structure based on a deep dense-residual network to apply loss on the generated reconstructed depth maps. Their approach of reusing the hierarchical features addresses the problem of blurred HR depth maps by exploiting features from different levels. This pyramid-structured network can progressively generate and apply loss on the reconstructed depth maps of various levels as a deep supervision scheme to reduce training difficulty and improve performance. Wei et al. \cite{wei2020component} proposed a component divide-and-conquer (CDC) model and a gradient-weighted loss, stacking three component-attentive blocks (CABs) to learn attentive masks and SR predictions. Their CDC model has a base model of an hourglass super-resolution network (HGSR), with a stacked hourglass architecture as an encoder-decoder and skip connection preserving spatial information at each resolution to predict pixel-wise outputs. This CDC model builds three CABs incorporating flats, edges and corners separately, while each CAB branch composes an intermediate SR and an attentive mask in the deep supervision layer. This CDC model is cross-tested on RealSR and their novel DRealSR datasets against seven existing methods, achieving outstanding SSIM results all above 0.800. Li et al. \cite{li2020pushing} argued that relatively soft labels are more suitable for SR deep supervision networks, therefore they proposed pushing and bounding loss to force the intermediate results from reconstructed internal features to get closer to ground truth, in order to achieve better final results. By utilising pushing and bounding loss to directly supervise all internal layers without explicit intermediate labels, and attained improvements on original models, outperforming RCAN baseline on Set5 \cite{bevilacqua2012low} and Set14 \cite{zeyde2010single} datasets with SSIM of 0.8960 and 0.7833 respectively for scale factor $\times 4$.

In SR, output images are highly correlated to input images and any recursive layer needs to maintain the exact copy of input values for recursions until the end of the network. Although typically deep supervision is utilised to optimise earlier layers in DNNs to address the gradient vanishing challenge, in SR networks, labels are either completely LR or completely high-resolution (HR), while mediately HR labels for deep supervision are difficult to be defined. Introducing DBDS networks to the SR task allows the model to supervise the internal features at different depths, addressing the gradient vanishing or explosion problem, and also regularising the network to converge towards a better final result. By adding the skip connection to an SR deep supervision network, the spatial information of input values is preserved and carried until the end of the network to predict pixel-wise outputs in the combination of the outputs generated from all the deeply supervised layers to yield natural SR output.

\begin{table}[ht]
\centering
{\scriptsize
\caption{Deep Supervision Embedded Super-Resolution Works}
\begin{tabular}{c|l|l|l}
\hline
Category & Label & Loss Sharing & Works \\
\hline
\hline
\multirow{3}{*}{DBDS} & \multirow{2}{*}{Same label} & Pro rata & \cite{8853318} \\
& & Dynamic pro rata & \cite{kim2016deeply} \\
& \multirow{2}{*}{Different labels} & Pro rata & \cite{li2020pushing} \\
& & Dynamic pro rata & \cite{wei2020component} \\
\hline
\end{tabular}
\label{tab:image-classification}
}
\end{table}
\section{Conclusion and Future Directions}
Based on the different ways deep supervision can be embedded into neural networks, we proposed three categories of deep supervision, i.e. hidden layers deep supervision (HLDS), different branches deep supervision (DBDS) and deep supervision post encoding (DSPE). HLDS is the simplest method, which can be applied in different applications to smooth the gradient propagation and increase the convergence speed during the learning process. DBDS is also applicable in various computer vision applications. Deep supervision modules in DBDS are mainly employed at different branches, representing different scale or depth level information. Such information is integrated together to make the final prediction, meanwhile, the output from different branches' deep supervision improves the transparency of the learning process. DSPE has mainly been used in heatmaps regression-based applications such as image segmentation and keypoint detection. This is because the heatmap-based supervision's output has a similar shape to the feature map propagating in the network so that such supervision's output can be easily encoded back to the main network, either for adding important features or for creating the attention map.

We conclude that deep supervision generally helps improve neural network performance in different computer vision tasks due to the following four aspects.
\begin{itemize}
\item First, deep supervision to some extent solves the gradient vanishing problem.
\item Second, deep supervision employed in the hidden layer can learn more discriminative features, which can contribute to the main network in different ways to improve performance.
\item Third, different results from deep supervision in a neural network can be integrated together to improve the final performance.
\item Fourth, deep supervision results in the neural network can to some extent make the neural network become more transparent.
\end{itemize}

Although most previous works show that deep supervision improves performance in various applications, deep supervision also has two limitations. 
\begin{itemize}
\item First, deep supervision in neural networks improves the backward propagation calculation complexity.
\item Second, deep supervision to some extent can cause overfitting problems due to including the extra loss to the main loss function. When optimising the main loss function, each loss led by deep supervision is optimised, which may cause the overfitting issue.
\end{itemize}

Based on the current knowledge gaps, we propose two future directions in the research area of deep supervision. First, it has been noted that some strategies in deep supervision works update different loss together at some early epochs and then freeze some shallow layer loss. These strategies prove to be effective in some applications, however, the optimal place and time to freeze the intermediate loss during learning remain unknown. Second, although deep supervision strengthens the directness of the model learning process, however, it may lead to overfitting problem.

The field of deep supervision is fast-moving and leads to many important real-world applications. There have been great advances in the last decade and it is likely, with the widespread use of technologies in homes and workplaces, that these developments will continue into the near future. The applications are widespread including medical, security, smartphone, traffic, etc, but there still remains some key research areas that require further work. This review has not only summarised the key technologies used and will be an important resource for computer scientists, but it also highlights the knowledge gaps where work will need to focus in the coming years.


%

\section*{Acknowledgment}




%

\bibliographystyle{unsrt}  
\bibliography{references}

\end{document}